\documentclass[10pt,twocolumn,letterpaper]{article}

\usepackage[pagenumbers]{cvpr} 

\usepackage{utfsym}
\usepackage{multirow}
\usepackage{booktabs}
\usepackage{float}
\usepackage{stfloats}

\usepackage{svg}
\usepackage{booktabs}

\newcommand{\algoname}{\textsc{EquiGen}\xspace}

\definecolor{cvprblue}{rgb}{0.21,0.49,0.74}
\usepackage[pagebackref,breaklinks,colorlinks,allcolors=cvprblue]{hyperref}

\def\paperID{4536} 
\def\confName{CVPR}
\def\confYear{2025}

\begin{document}
\title{Few-shot Implicit Function Generation via Equivariance}


\author{Suizhi Huang$^{1,2}$, \, Xingyi Yang$^{2}$, \, Hongtao Lu$^{1}$\footnotemark[1], \, Xinchao Wang$^{2}$\footnotemark[1]\\
$^1$ Shanghai Jiao Tong University \quad
$^2$ National University of Singapore\\
}


\setlength{\abovedisplayskip}{3pt} 
\setlength{\belowdisplayskip}{3pt}
\maketitle

\renewcommand{\thefootnote}{\fnsymbol{footnote}}
\footnotetext[1]{Corresponding author.}

\begin{abstract}


Implicit Neural Representations (INRs) have emerged as a powerful framework for representing continuous signals. However, generating diverse INR weights remains challenging due to limited training data. We introduce \textbf{Few-shot Implicit Function Generation}, a new problem setup that aims to generate diverse yet functionally consistent INR weights from only a few examples. This is challenging because even for the same signal, the optimal INRs can vary significantly depending on their initializations. To tackle this, we propose \algoname, a framework that can generate new INRs from limited data. The core idea is that functionally similar networks can be transformed into one another through weight permutations, forming an equivariance group. By projecting these weights into an equivariant latent space, we enable diverse generation within these groups, even with few examples. \algoname implements this through an equivariant encoder trained via contrastive learning and smooth augmentation, an equivariance-guided diffusion process, and controlled perturbations in the equivariant subspace. Experiments on 2D image and 3D shape INR datasets demonstrate that our approach effectively generates diverse INR weights while preserving their functional properties in few-shot scenarios.


\end{abstract}
\section{Introduction}
\label{sec:intro}



Open-sourced models have been the driving force behind the incredible progress of Artificial Intelligence~(AI)~\cite{alphafold,clip,llama}. Beyond just making powerful models accessible, this openness has created new opportunities for meta-learning: treating model weights themselves as data sources. Recent works have demonstrated the potential of learning from~\cite{equivariant,signal,neumeta,inr2vec,herrmannlearning} and generating neural network weights~\cite{Peebles2022,knyazev2021parameter,functa,hypertransformer} to promote further research.

One type of neural network well-suited for the weight generation paradigm is the Implicit Neural Representation (INR). These networks use a simple multi-layer perceptron (MLP) to fit continuous signals~\cite{siren, inr2, inr3}. INRs serve as powerful tools for representing continuous data, offering several advantages: high-fidelity reconstruction~\cite{deepsdf,nerf}, smooth interpolation properties~\cite{inr3,mnif}, and infinite output resolution~\cite{mipnerf}. Their consistent and comparative simple architecture makes them ideal candidates for weight-based generative modeling.

\begin{figure}[tbp]
    \centering
    \includegraphics[width=1\linewidth]{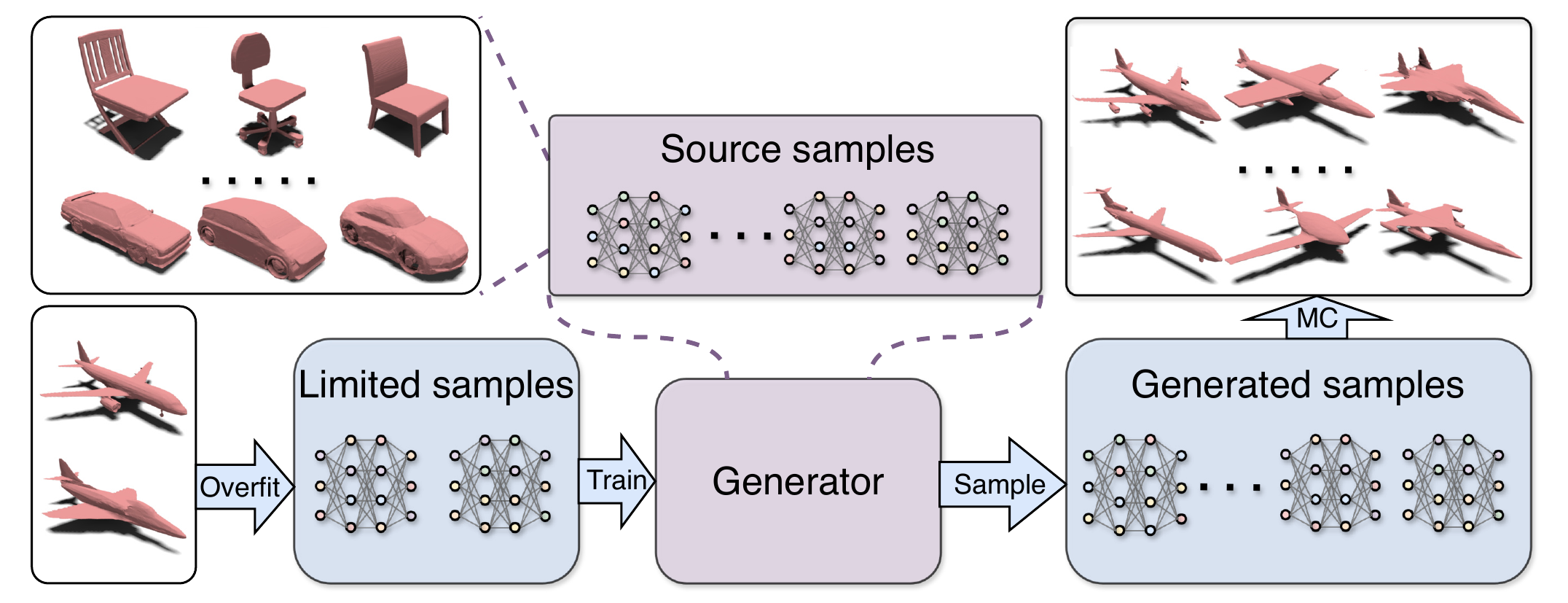}
    \caption{Illustration of the Few-shot Implicit Function Generation setting with 3D INR data examples. The goal is to generate diverse INR weights from limited target samples. Source samples (top) show previously observed INRs of diverse shape categories. In practice, only limited target samples (bottom left) are available for training. The framework aims to learn a generator that can produce diverse generated samples (right) despite the limited training data. This setting addresses the practical scenario where only a few examples of new shapes are available for training.} 
    \label{fig:intro}
    \vspace{-5mm}
\end{figure}

However, training generative model on weight comes with its challenges, primarily due to a shortage of data. Currently, collecting large-scale INR data is costly, as it involves time-consuming gradient-based optimization. Existing datasets for INR data are limited from the diversity of sources~\cite{implicitzoo,equivariant}. This data gap imposes restrictions on further investigation of weight space.

Given these constraints, an intuitive motivation is to target on a more practical task of \textbf{Few-shot Implicit Function Generation}. As illustrated in ~\cref{fig:intro}, with only a few new INR checkpoints, we aim to train a generative model that can produce diverse weights while preserving the original distribution's function of these few INR data. These generated checkpoints serve multiple purposes: they can be utilized as new data sources for weight space learning, or they can be directly applicable for practical downstream tasks like 3D shape generation.

This new problem comes with two fundamental challenges. First, the curse of dimensionality~\cite{bellman1966dynamic} exacerbates data demands. Since model weights reside in high-dimensional spaces, a limited number of samples cannot adequately capture the full weight distribution, resulting in poor generalization. Second, conventional few-shot learning methods assume that similar samples have element-wise similarity~\cite{fewalign,anomaly}. However, this assumption doesn’t hold for INR checkpoints. INR data can vary significantly in individual weights yet still yield similar input-output behavior, especially under weight permutations\cite{nern,spectral}. This structural difference makes it difficult to apply standard knowledge transfer techniques effectively, results in potential mode collapse. 


Previous approaches to parameter generation, while potentially applicable to Few-shot Implicit Function Generation, face significant limitations and fail to address these fundamental challenges. Some methods use generative models to directly reconstruct the weight distributions~\cite{hyperdiffusion,Peebles2022}, they struggle with the curse of dimensionality. While some other methods attempt to address this through autoencoder-based dimensionality reduction~\cite{inr2vec,schurholt2022hyper,wang2024neural}, they fail to respect the inherent symmetry of the weight space, leading to poor generalization performance.

To address these challenges, we resort to the principle of \textbf{equivariance}, \textit{i.e.}, weight space symmetry~\cite{equivariant,deepset,deepset2}. It means that, for arbitrary neural networks, permuting output channels in one layer and reversing this permutation in the next layer preserves the network’s functionality. Thus, all weights that result from such permutations form an equivariance group~\cite{equivariant}. 
Leveraging this property, we aim to design a generative model capable of producing all samples within an equivariance group from a few checkpoints. In this way, we significantly reduce the data requirements in the few-shot setting. 

Based on that, we design a framework named \algoname. Our key insight of equivariance in few-shot setting is that, by projecting weights into an equivariant latent space, we could achieve diverse generation inside an equivariance group from limited examples. 

The complete framework contains three stages.

\begin{itemize}
    \item \textbf{Equivariant Encoder.} We design an encoder that projects weights into an equivariant latent space. Using contrastive learning with smooth augmentations, the encoder maps functionally equivalent weights to similar representations. This establishes a structured latent space that captures the concept of equivariance.
    \item \textbf{Equivariance-Guided Diffusion.} We develop a diffusion process guided by equivariance. The denoising steps are conditioned on previous equivariant features and are regularized using an equivariant loss, ensuring that the generated weights retain both distributional properties and equivariance.
    \item \textbf{Controlled Perturbations for Diversity.} To address the challenge of generating diverse yet consistent weights, we apply controlled perturbations in the equivariant subspace. These perturbations enable us to explore the full equivariance group of each example, generating weights that are both diverse and functionally consistent.
\end{itemize}

We evaluate \algoname on various INR data including 2D images and 3D shapes. Our findings reveal that \algoname not only boosts high quality generation, but also encourages diversity. Ablation studies support the effectiveness of each designed module, emperically prove the importance of equivariance.

Our contributions are summarized as follows:
\begin{itemize}
    \item We introduce a practical setting named \textbf{Few-shot Implicit Function Generation} with an equivariance-based framework \algoname that respect to weight equivariance to enable diverse yet functionally consistent weight generation from limited samples.
        
    \item We present three key innovations that systematically exploit equivariance: 1) An equivariant encoder learned through a contrastive learning paradigm with smooth augmentation for robust equivariant feature learning; 2) Equivariance-guided diffusion with explicit equivariance regularization; 3) Controlled equivariant subspace perturbation for diverse generation.
    
    \item We demonstrate the effectiveness of our \algoname through comprehensive experiments on INR data from both 2D images and 3D shapes.
\end{itemize}
\section{Related Work}
\label{sec:related work}

\subsection{Implicit Neural Representation}
Implicit Neural Representations (INRs) have demonstrated remarkable efficacy in representing diverse forms of complex signals, including spatial occupancy~\cite{occ1,wire,siren}, 3D geometric morphology~\cite{chen2019learning,chabra2020deep,genova2019learning}, signed distance functions~\cite{deepsdf,dist}, 3D scene appearence~\cite{xie2022neural,chen2023neurbf,muller2022instant,ma2023deformable,li2023dynibar} and some other complex signals~\cite{mcginnis2023single,xu2023nesvor,ma2024continuous,inr3} with the help of a small neural network, usually a MLP with few layers. 


\subsection{Equivariant Architectures}
A wide range of studies have tried to build equivariant architecture to respect underlying symmetries in the data. This leads to advantages including smaller parameter space, efficient implementation, and better generalization abilities~\cite{sphericalcnn,generalizationofequi,equivariant,zhou2024neural}. The standard construction pattern involves identifying basic equivariant functions which are often linear~\cite{deepset,deepset2,deepset3}, and composing them with pointwise nonlinearities to build deep networks. Of particular relevance to our work are architectures designed for set-structured data, where the input represents ordered elements requiring equivariance to permutations. A recent study incorporates advantages of several previous works to build an efficient and highly expressive architecture~\cite{equivariant}. These foundations prove crucial for our work, as neural network weight spaces naturally exhibit similar symmetry structures.


\subsection{Neural Network Weight Generation}

Recently, neural network weight generation has gained significant attention, from early meta-learning approaches~\cite{hypernetworks,maml} to recent generative modeling techniques. Existing methods broadly split into two categories: direct weight space diffusion methods~\cite{hyperdiffusion,Peebles2022}, which struggle with high-dimensional spaces, and dimensionality reduction approaches using autoencoders~\cite{inr2vec,schurholt2022hyper,wang2024neural}, which often fail to preserve the inherent structural characteristics of weight spaces. The key distinction of our method lies in its systematic implementation of equivariance, which aims at preserving these inherent characteristics.


\begin{figure*}[htbp]
    \centering
    \includegraphics[width=1.0\linewidth]{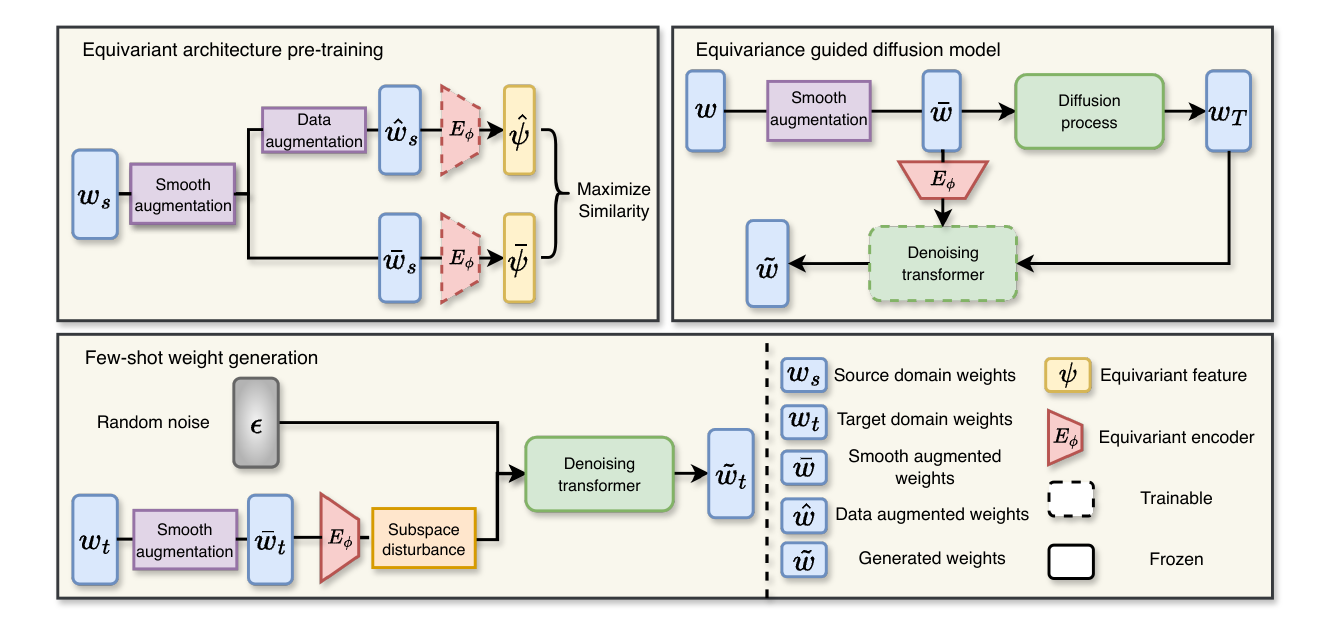}
    \vspace{-8mm}
    \caption{Overview of our \algoname framework. The method consists of three stages: (1) Equivariant Encoder Pre-training through contrastive learning with smooth and INR-based augmentations, (2) Distribution Modeling via a diffusion process conditioned on learned equivariant features, and (3) Few-shot Adaptation using equivariant subspace disturbance for diverse weight generation. Our framework leverages the inherent equivariance of neural network weights to address both generalization and mode collapse challenges.}
    \label{fig:arch}
    \vspace{-6mm}

\end{figure*}

\section{Preliminaries}
In this section, we formalize the \textbf{Few-shot Implicit Function Generation} problem, we also interpret fundamental properties of equivariance and corresponding equivariant architectures.

\subsection{Problem Definition} \label{sec:def}
Consider a class space $\mathcal{F}$ where each $f \in \mathcal{F}$ represents a category of signals (\textit{e.g.}, images of a specific digit or 3D shapes from a particular object class such as cars or planes) that can be encoded by implicit neural representations. Let $\mathcal{W}$ denotes the weight space of MLPs with a fixed architecture, and $\Phi: \mathcal{W} \rightarrow (\mathcal{X} \rightarrow \mathcal{Y})$ maps weights to their corresponding implicit functions.

\begin{definition}[Few-shot INR Generation]
\vspace{-2mm}
Given $k$ example weights $S_f = \{w_1,...,w_k\}$ that encode valid instances of a target class $f$, the objective is to generate new weights that encode diverse yet valid instances of the same class. Formally, we aim to learn a generator $G: S_f \mapsto \tilde{p}_f(w)$ that estimates the weight distribution of class $f$, where validity requires the reconstructed signal $\Phi(w)$ to maintain class-specific properties while allowing intra-class variation.
\vspace{-2mm}
\end{definition}

A comprehensive mathematical formalization is provided in the Appendix. 

\subsection{Equivariance and Equivariant architecture}
\label{sec:pre2}

Previous studies have revealed that neural networks exhibit inherent equivariance in their weight space. Different weight configurations can represent the same function due to the permutation invariance of neurons within each layer~\cite{neumeta,symmetry1,symmetry2,deepset3}. Understanding and leveraging this equivariance is crucial for Few-shot Implicit Function Generation, as it enables more efficient learning from limited examples by exploiting the underlying structure of the weight space~\cite{sphericalcnn,generalizationofequi,so3,permutationequi}. Since INR encoding networks typically employ MLPs that inherently possess this property, we present key concepts of it and corresponding equivariant architectures.

\noindent\textbf{Equivariance.}
Consider a shallow network with weight matrices $W_1$ and $W_2$. For any permutation matrix $P$, the transformed weights $PW_1$ and $W_2P^T$ yield a functionally equivalent network. This symmetry emerges from the equivariance of pointwise activation functions:
\begin{equation}
P\sigma(x) = \sigma(Px).
\end{equation}
This property implies that the space of functionally equivalent networks forms orbits under permutation groups, effectively reducing the sample complexity required for learning.

\noindent\textbf{Equivariant Architectures.}
To exploit this equivariance, we construct architectures that respect the inherent structure of the weight space. A function $L:\mathcal{V} \rightarrow \mathcal{W}$ is \textit{equivariant} if it commutes with group actions $\rho_1$ and $\rho_2$ for all $v \in \mathcal{V}, g \in \mathcal{G}$:
\begin{equation}
L(\rho_1(g)v) = \rho_2(g)L(v),
\end{equation}
where $\mathcal{V}$ and $\mathcal{W}$ are vector spaces, and $\mathcal{G}$ is a group.
Following previous works~\cite{deepset,deepset2,deepset3}, we define the canonical form of an equivariant encoder:
\begin{equation}
F_{\text{equi}}(x) = L_k \circ \sigma \circ \dots \circ \sigma \circ L_1(x),
\end{equation}
where each $L_i$ is an equivariant affine transformation and $\sigma$ is a pointwise activation function. By incorporating these symmetries into our architecture, we significantly enhance sample efficiency in few-shot learning. The generator learns to explore the entire orbit of functionally equivalent weights from a few examples, effectively amplifying the limited training data.

\section{Methodology}

\subsection{Overview}
Follow the definition of Few-shot Implicit Function Generation problem, our goal is to learn how to generate weights that follow a specific distribution, using only a small number of samples as reference. This problem presents two fundamental challenges as stated in \cref{sec:intro}: (1) poor generalization due to the high dimensionality of weight space, and (2) ineffective knowledge transfer and potential mode collapse due to weak element-wise similarity between functionally equivalent weights. We address both challenges through \textbf{equivariance} which is interpreted in \cref{sec:pre2}.

Our key insight is that by projecting weights into an equivariant latent space, we can generate diverse samples within an equivariance group using only a few reference examples inside this latent space. This projection effectively reduces dimensionality while preserving functional relationships, enabling both better generalization and meaningful knowledge transfer. As illustrated in \cref{fig:arch}, we systematically leverage equivariance across three stages:

\noindent\textbf{Equivariant Feature Learning:} 
We establish a foundation through an equivariant encoder trained via contrastive learning with weight space smooth augmentations, ensuring learned representations preserve essential equivariance property while remaining expressive.

\noindent\textbf{Equivariance-Guided Distribution Modeling:} 
A diffusion-based generator, conditioned on these equivariant features, models the weight distribution while preserving symmetry properties through a designed explicit equivariance regularization.

\noindent\textbf{Equivariant Few-shot Adaptation:} 
For k-shot generation, we introduce controlled disturbances in the equivariant subspace, enabling diverse yet functionally consistent generation by exploiting the inherent symmetry structure from the equivariant features.

\subsection{Equivariant Encoder Pre-training}
A compact and expressive latent space of equivariant features is crucial for effective distribution modeling, improving diffusion convergence and stability. To learn such features, we employ a contrastive learning framework specifically adapted for our equivariant architecture. Additionally, we introduce a weight-specific smooth augmentation strategy to enhance the robustness and representation power of the equivariant encoder.

\noindent\textbf{Equivariant Architecture.}
Similar to previous work, our equivariant encoder, inspired by the symmetry structure of deep weight spaces, operates directly on weight matrices through a carefully designed architecture that preserves functional equivariance~\cite{deepset,equivariant}. The encoder decomposes the input weight space into meaningful sub-representations corresponding to different components of a neural network and implements equivariant mappings between these sub-representations through a combination of linear equivariant layers, invariant layers, and efficient pooling and broadcasting operations.
Formally, the encoder can be denoted as a mapping $E_\phi: \mathcal{V}\rightarrow\mathcal{V}$ parameterized by $\phi$ following the canonical form demonstrated in \cref{sec:pre2}.
This architecture enables effective projection of the input weight space into an expressive low-dimensional equivariant subspace while preserving the underlying functional relationships. Detailed architecture design can be found in Appendix.

\noindent\textbf{Smooth Augmentation.}
The key idea of this encoder is to map weights from an equivariance group to a compact cluster in the equivariant subspace. Yet it can indeed learn from unconstrained weight spaces, recent studies reveal that modern neural networks fundamentally rely on smooth signal modeling~\cite{spectral,neumeta}. Consequently, to enhance our encoder's feature learning capability, we introduce a smooth augmentation technique for the input weight space. This augmentation could be seen as an optimization of the starting point, and could lead to better representation power of the pre-trained encoder.

Specifically, we represent neural networks as dependency graphs $G = (V,E)$~\cite{depgraph}, where nodes represent operations with weights and edges denote inter-connectivity. For each subgraph, we find a permutation matrix $P$ that minimizes the total variation ($TV$) across the network~\cite{totalvariation}.
Leveraging the orthogonal independence of permutations~\cite{neumeta}, we decompose this optimization into multiple Shortest Hamiltonian Path problems~\cite{hamilton}, which we efficiently solve using a 2.5-opt local search algorithm~\cite{2.5optTSP}. The resulting optimal permutation $P^*$ is applied to all weight matrices within each network subdivision, ensuring smooth weight space augmentation while preserving functional equivariance.

\begin{figure}[tbp]
    \centering
    \includegraphics[width=1\linewidth]{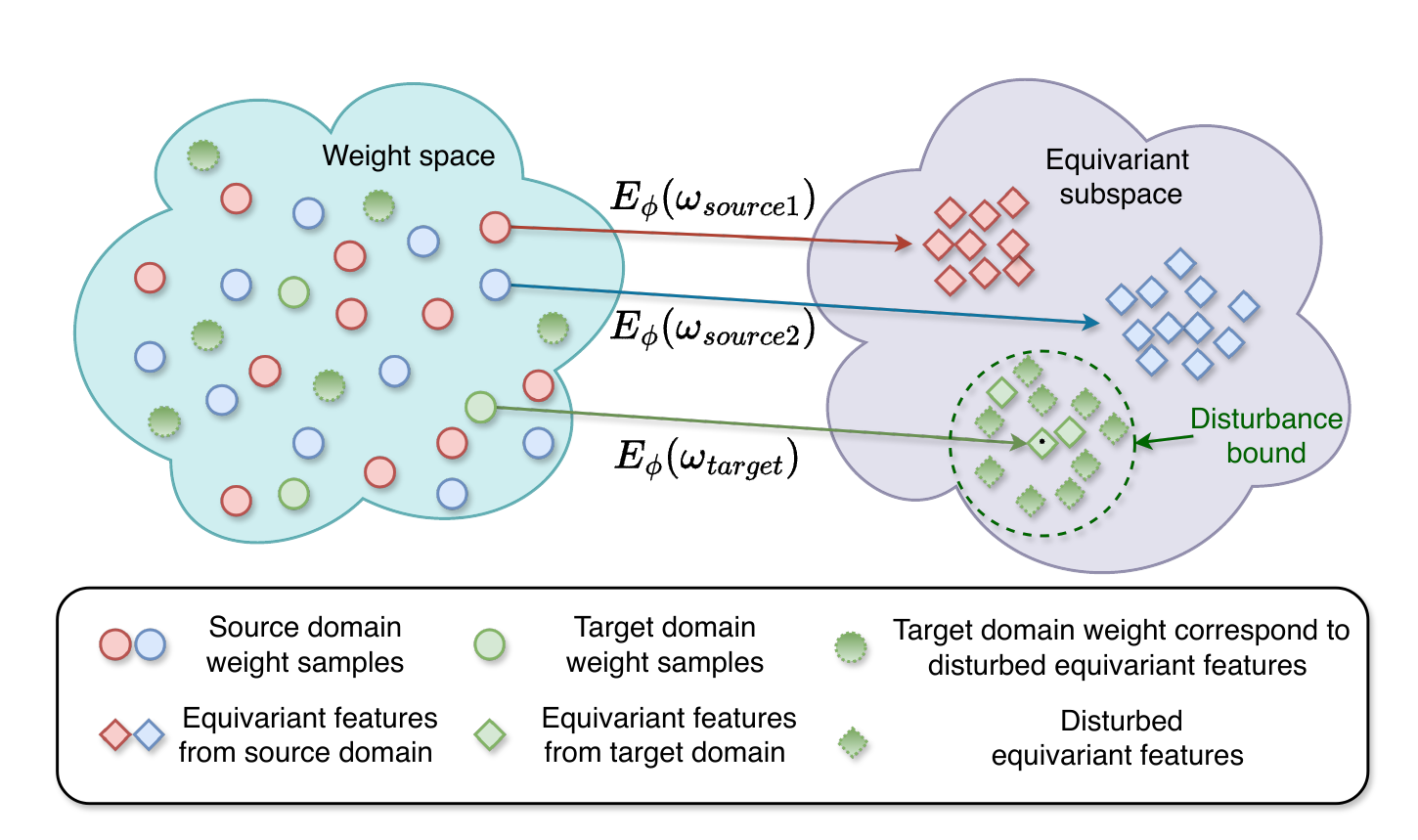}
    \vspace{-6mm}
    \caption{Equivariant architecture $E_\phi$ aims to map weights from the same equivariance group to similar representations, creating a structured latent space that captures the inherent symmetries of neural networks. By leveraging this equivariant subspace, we can implement a controlled disturbance strategy that sample diverse equivariant features while maintaining class consistency.}
     \vspace{-6mm}
    \label{fig:equi}
\end{figure}

\noindent\textbf{Contrastive Learning.}
Following the SimCLR framework~\cite{simclr}, we pre-train our equivariant encoder to maximize similarity between different augmented views of the same weight data in the equivariant latent space. The process involves three key steps:

\noindent 1) We generate positive pairs by applying stochastic INR-based augmentations to smoothed weights $\bar{w}$, obtaining augmented weights $\hat{w}$. These augmentations include rotation, translation, scaling, and our proposed color jittering and bias perturbation (see Appendix for details).


\noindent 2) Extract equivariant features through pre-trained equivariant encoder $E_\phi$: $\bar{\psi} = E_\phi(\bar{w})$ and $\hat{\psi} = E_\phi(\hat{w})$.

\noindent 3) Optimize using contrastive loss:
\begin{equation}
\ell_{i,j} = -\log\frac{\exp(sim(E_\phi(\check{w}_i), E_\phi(\check{w}_j)) / \tau)}{\sum_{k\neq i} \exp(sim(E_\phi(\check{w}_i), E_\phi(\check{w}_k)) / \tau)},
\end{equation}
where $sim(u,v)$ denotes cosine similarity, $\tau$ is a temperature parameter. $\{\check{w}_k\}$ contains both the smoothed weights $\{\bar{w}_k\}$ and the augmented weights $\{\hat{w}_k\}$.

\begin{figure}[t]
    \centering
    \includegraphics[width=1\linewidth]{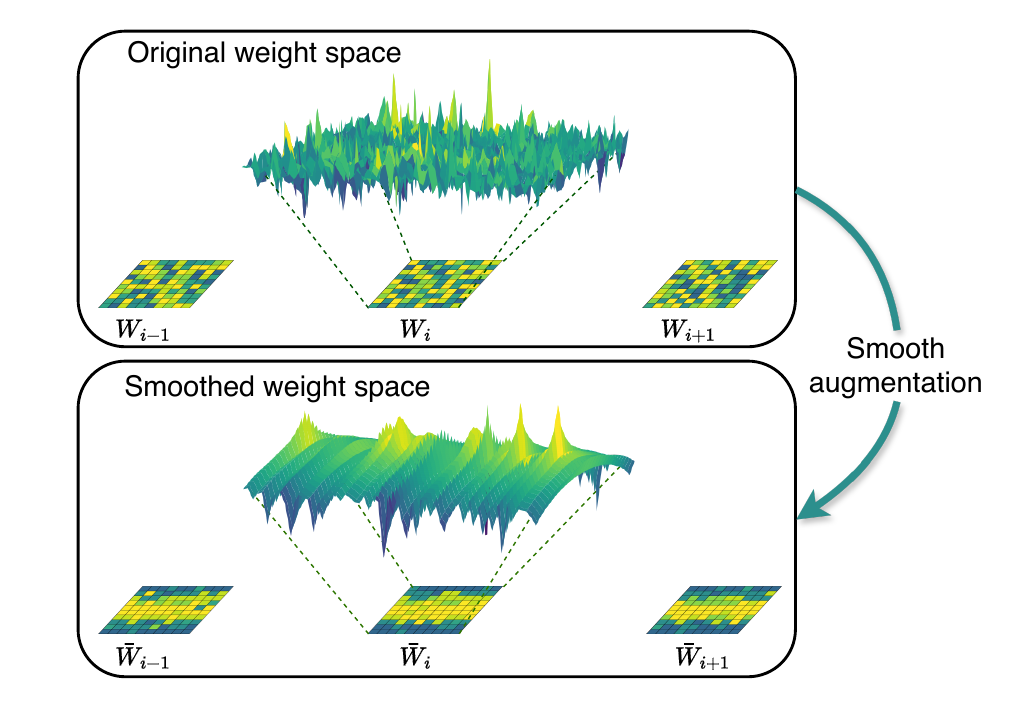}
    \vspace{-6mm}
    \caption{To optimize the equivariant feature extraction, we seek permutations that minimize the weight matrix's total variation. This smoothing operation reduces abrupt discontinuities in the weight space, facilitates more effective learning of inherent equivariant properties by starting from an optimized point. The smoother weight manifold (bottom) enables better equivariant feature capture compared to the original space (up).}
    \label{fig:smooth}
    \vspace{-6mm}
\end{figure}

\subsection{Equivariance Guided Diffusion}
\label{sec:diff}

Within this stage, we employ an equivariance-conditioned diffusion model to reconstruct the input weight distributions, utilizing their corresponding equivariant features. This approach enables a robust reconstruction of the weight space, while maintaining the equivariance inherent in the original data structure.

\noindent\textbf{Weight-Space Diffusion.}
The modeling begins by transforming each set of smoothed weights $\{\bar{w}_i\}$ into a flattened one-dimensional vector representation. Subsequently, these smoothed weights $\bar{w}_i$ undergo a diffusion process, yielding noised samples $\bar{w}_T$. For the denoising network, we implement a transformer architecture $G_\theta$. This choice leverages transformers' established capability to efficiently process long vector sequences, advantageous for weight space manipulation~\cite{Peebles2022,hyperdiffusion}. To highlight the difference with normal diffusion models, our denoising network predicts the denoised weights directly, rather than the noise~\cite{Peebles2022}. The architecture also incorporates cross-attention layers following each transformer block~\cite{ldm}, enabling the sampled weights $\tilde{w}_i$ to be conditioned on equivariant features $\psi_i$. These features are derived by processing the smoothed weights $\bar{w}_i$ through the former pre-trained equivariant encoder $E_\phi$. The comprehensive denoising procedure can be formally expressed as:
\begin{equation}
    \tilde{w}_i = G_\theta(\bar{w}_T, E_\phi(\bar{w}_i)).
\end{equation}
\noindent\textbf{Training and Optimization.}
The primary goal of our approach is to optimize the denoising network to model the input distribution accurately. We apply a simple Mean Squared Error (MSE) loss between the denoised weights $\tilde{w_i}$ and the smoothed weight $\bar{w_i}$ denoted as $\mathcal{L}_{recon}(\bar{w},\tilde{w})$.

However, in order to leverage the characteristic of equivariance in the following few-shot generation stage, we also designed a specific loss to regulate the approximation of the equivariant features of the generated weights and the original smoothed weights:
\begin{equation}
    \mathcal{L}_{eq}(\bar{w},\tilde{w}) = \frac{1}{N}\sum_{i=1}^{N} \|E_\phi(\bar{w}_i) - E_\phi(\tilde{w}_i)\|_2^2.
\end{equation}

By composing these two different objectives, our final target is to minimize the composite objective by sampling all the smoothed weight and generated weight:
\begin{equation}
    \min_{\theta} \mathbb{E}_{\bar{w}_i, \hat{w}_i} \left[ \mathcal{L}_{total} \right] = \min_{\theta} \mathbb{E}_{\bar{w}_i, \hat{w}_i} \left[ \mathcal{L}_{recon} + \lambda \mathcal{L}_{eq} \right].
\end{equation}
This loss function not only ensuring accurate distribution reconstruction, but also boosting the generalization.

\subsection{Few-shot Fine-tuning}
Through latent space guidance, the diffusion model demonstrates adaptability to previously unseen distributions, given only a limited set of weights $\{w'_1,\dots,w'_k\}$ from the target distribution. This adaptation is facilitated through the utilization of equivariant features $\{\psi'_1,\dots,\psi'_k\}$ extracted via the pre-trained equivariant encoder, serving as novel guidance vectors that enable knowledge transfer. The adaptation process need only a few iterations of fine-tuning, following the same procedures in \cref{sec:diff}.

\noindent\textbf{Subspace Disturbance.}
To enhance the diversity of generated weights, we introduce an equivariant subspace disturbance strategy during the final generation. This approach is motivated by the observation that the diffusion model's training process inherently embeds knowledge within the equivariant subspace through feature-guided denoising. Consequently, the diversity of the guiding equivariant features directly influences the variability of the generated weights, establishing a foundation for controlled diversity in weight generation.

Specifically, we apply a random Gaussian noise $\epsilon\sim\mathcal{N}(0,I)$ on the equivariant feature $\psi'_i$ with a controlling parameter $\gamma$: $\tilde{\psi}'_i = \psi'_i + \gamma \epsilon$. We impose this constraint on the magnitude of the disturbance to make it bounded and controllable. Finally, the generation of target weights is accomplished through a denoising procedure applied to $k$ independent Gaussian noise vectors $\mathcal{E} = \{\epsilon_1, \dots, \epsilon_k\}$. This process is guided by disturbed equivariant features $\{\tilde{\psi}'_1,\dots,\tilde{\psi}'_k\}$ derived from the support set.

\section{Experiments}


\subsection{Experimental Setup}
We establish two Few-shot Implicit Function Generation scenarios: 1) Given a set of MLP encoded INRs of different \textbf{2D images}, we split them into two disjoint parts: the seen categories $S_{source}$ and the unseen categories $S_{test}$, where category is defined by their rendered image. 2) Similarly, given a set of MLP encoded INRs of different \textbf{3D shapes}, which are split into two disjoint parts: the seen categories $S_{source}$ and the unseen categories $S_{test}$, where category is defined by their rendered 3D shape.

\noindent\textbf{Datasets.}
For the 2D image scenario, we evaluate our methodology on two benchmark datasets: MNIST (2D greyscale images)~\cite{mnist} and CIFAR-10 (2D RGB images)~\cite{cifar10}. We utilize the pre-overfitted INR dataset for MNIST (MNIST-INRs) provided by~\cite{equivariant} and the corresponding dataset for CIFAR-10 (CIFAR-10-INRs) from~\cite{implicitzoo}. MNIST-INRs encompasses 50K INR instances, with 5K samples per category, while CIFAR-10-INRs contains 60K INR instances, comprising 6K samples per category. For the 3D shape domain, we evaluate our approach on the ShapeNet dataset~\cite{shapenet}, focusing on three representative categories: airplane (4045 shapes), car (6778 shapes) and chair (3533 shapes). These shapes are encoded into MLPs, denoted as ShapeNet-INRs.
Regarding architectural specifications, CIFAR-10-INRs employs a 3-layer SIREN MLP with a width of 64~\cite{siren}, while MNIST-INRs utilizes a 3-layer SIREN MLP with a width of 32~\cite{siren}. For ShapeNet-INRs, it implements a 3-layer standard MLP with a width of 128, incorporating ReLU activation functions and input positional encoding~\cite{hyperdiffusion}.

In the context of few-shot weight generation, we partition each dataset into two distinct subsets~\cite{few1,few2,myspot,weditgan}. The unseen category support set $S_{test}$ comprises randomly sampled INR data from a single category, with the sample size varying from 1 to 10 to evaluate different levels of difficulty. The remaining data from other categories constitute the seen category support set $S_{source}$.


\noindent\textbf{Implementation}
We provide important configurations here, please refer to Appendix for more detailed information. Our equivariant architecture is implemented with four hidden equivariant layers~\cite{equivariant}, with the output equivariant feature dimension set to 128. Across the three stages: In the equivariance-guided diffusion stage, we utilize a squared cosine beta scheduler~\cite{improveddiffusion} across 1000 timesteps and implement DDIM~\cite{ddim} for sampling. The equivariance loss proportion parameter $\lambda$ is set to 0.1 unless otherwise specified.
Finally, during the generation process, the subspace disturbance parameter $\gamma$, which controls the noise intensity, is set to 0.3 by default.


\noindent\textbf{Metrics.}
Since there is no direct evaluation metrics for generated weights, we adopt an indirect evaluation strategy aligned with existing works~\cite{functa,inr2vec,wang2024neural,schurholt2022hyper,hyperdiffusion}. For 2d images, a direct rendering is applied and for 3d shape scenario, we extract the underlying isosurface from the MLPs with Marching Cubes~\cite{mc}. Our evaluation framework assesses both the quality and diversity of generated results for both two scenarios:
For the 2D image scenario, we employ FID~\cite{fid} and LPIPS~\cite{lpips} as the metrics. Specifically, we apply the intra-cluster version of LPIPS to quantify the diversity of generated unseen images in our few-shot generation context like many other few-shot image generation settings do~\cite{weditgan,myspot,ldm}. Lower FID indicates better quality and higher LPIPS indicates enhanced diversity.
For the 3D shape scenario, we follow prior works~\cite{dpc,lion,pvd,meshgpt} in evaluating MMD, COV, and 1-NNA. For MMD, lower is better; for COV, higher is better; for 1-NNA, 50\% is the optimal. We compute these metrics using Chamfer Distance (CD) as the underlying distance measure, with reported CD values scaled by a factor of $10^2$ for clarity.

\begin{table}[]
\caption{Quantitative comparison of 10-shot generation performance on ShapeNet-INRs, where each test set consists of INRs from a single unseen object class. \textbf{Bold} indicates the best result and \underline{underline} indicates the second best result.}
\vspace{-3mm}
\label{fig:3d}
\resizebox{\columnwidth}{!}{%
\begin{tabular}{c|c|ccc}
\toprule
Category          & Method         & \multicolumn{1}{c}{MMD$\downarrow$} & \multicolumn{1}{c}{COV(\%)$\uparrow$} & \multicolumn{1}{c}{1-NNA(\%)$\downarrow$} \\ \midrule
\multirow{6}{*}{Airplane}  
 & PVD~\cite{pvd}            &  5.6  &    26  &  84.7\\
 & DPC ~\cite{dpc}           & \underline{4.1}  &   29   &   84.6 \\
  & INR2Vec~\cite{inr2vec} &   5.1   &  29    &  80.9    \\
  &Voxel Baseline~\cite{hyperdiffusion} &   8.5   &  20  &    98.3  \\
 & HyperDiffusion~\cite{hyperdiffusion} &         5.0            &        \underline{33}          & \underline{80.6}               \\
 & \textbf{\algoname (Ours)}  &    \textbf{3.4}        &      \textbf{35}    &    \textbf{73.0}       \\ \midrule
\multirow{6}{*}{Car}      
 & PVD~\cite{pvd}            & 4.7  &  \underline{24}  &  \underline{82.1} \\
 & DPC~\cite{dpc}            &  \underline{4.2}        &    20      &   85.3        \\
  & INR2Vec~\cite{inr2vec} &   4.8   &   19       &   87.0   \\
  & Voxel Baseline~\cite{hyperdiffusion} &  5.8 & 11  &   97.9    \\
 & HyperDiffusion~\cite{hyperdiffusion} &   4.4     &       22     &      83.4     \\
 & \textbf{\algoname (Ours)}  &    \textbf{3.5}      &        \textbf{31}     &     \textbf{76.5}     \\ \midrule
\multirow{6}{*}{Chair}    
 & PVD~\cite{pvd}            &  7.1   &  21 &  80.0   \\
 & DPC~\cite{dpc}            &  \underline{6.5}  &   23   &   81.5            \\
  & INR2Vec~\cite{inr2vec} &   6.8    &  25   &    \underline{71.7}    \\
  & Voxel Baseline~\cite{hyperdiffusion} & 12.4    & 13    & 88.5 \\
 & HyperDiffusion~\cite{hyperdiffusion} &     7.2     &    \underline{28}     &   73.1    \\
 & \textbf{\algoname (Ours)}  &        \textbf{4.2}            &       \textbf{41}              &    \textbf{67.1}              \\ \bottomrule
\end{tabular}%
}
\end{table}
\begin{table}[]
\caption{Quantitative comparison of 10-shot generation performance on MNIST-INRs and CIFAR-10-INRs.}
\vspace{-3mm}
\label{fig:2d}
\centering
\footnotesize
\begin{tabular}{c|c|cc}
\toprule
 Dataset                & Method        & FID$\downarrow$      & LPIPS$\uparrow$     \\ \midrule
\multirow{6}{*}{MNIST-INRs} & FIGR~\cite{figr}         &       160.26               &   0.1399                              \\
                       & DAGAN~\cite{dagan}         &       149.77              & 0.2001                             \\
                       & DAWSON~\cite{dawson}        &      164.90          &    0.1830                   \\
                       & INR2Vec~\cite{inr2vec}        &  \underline{157.20}  &   0.2185              \\
                       
                       & HyperDiffsion~\cite{hyperdiffusion}        &  165.91      &       \underline{0.2727}       \\
                       &\textbf{\algoname (Ours)} &   \textbf{121.24}      & \textbf{0.4133}              \\ \midrule
\multirow{6}{*}{CIFAR10-INRs} & FIGR~\cite{figr} & 225.05 & 0.1728  \\
                       & DAGAN~\cite{dagan}      & 200.11 & 0.2250  \\
                       & DAWSON~\cite{dawson}        &  204.62  &   0.2191                  \\
                       & INR2Vec~\cite{inr2vec}        & 188.39  &  0.2712            \\
                       & Hiperdiffusion~\cite{hyperdiffusion}        &  \underline{186.58}   &   \underline{0.3104}        \\
                       & \textbf{\algoname (Ours)} & \textbf{164.14} &\textbf{0.4926}   \\ \bottomrule
\end{tabular}%
\vspace{-5mm}
\end{table}

\noindent\textbf{Baselines.}
Our method is benchmarked against two different types of methods: modality-based methods and INR-based methods.
In the first category, we benchmark against methods that directly manipulate domain-specific representations. For the 2D image domain, we compare our approach with representative few-shot image generation methods, including FIGR~\cite{figr}, DAGAN~\cite{dagan} and DAWSON~\cite{dawson}. Similarly, in the 3D shape domain, we evaluate against prominent 3D shape generation methods such as PVD~\cite{pvd} and DPC~\cite{dpc}.
In the second category, we evaluate against methods that operate directly on INR representations rather than domain-specific data formats. Specifically, we compare our approach with two representative frameworks: INR2Vec~\cite{inr2vec} and HyperDiffusion~\cite{hyperdiffusion}, both of which are assessed across 2D image and 3D shape scenarios to ensure comprehensive evaluation. Specifically, HyperDiffusion has a voxel version (Voxel Baseline) whose INR data are designed to encode voxel information instead of meshes.

\begin{figure}[t]
    \centering
    \includegraphics[width=1\linewidth]{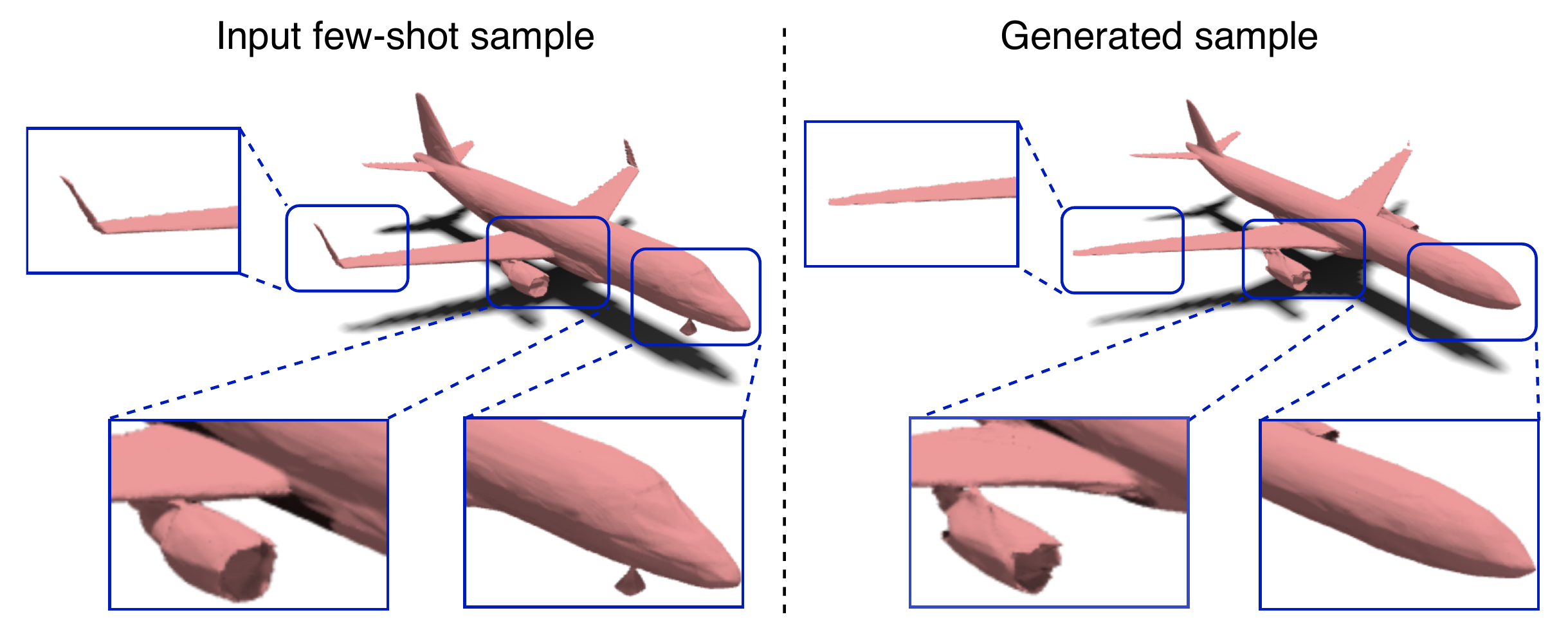}
    \vspace{-5mm}
    \caption{Visualizations of generated ShapeNet-INRs from the few-shot airplane example. The outputs exhibit diverse shape variations while preserving the airplane category characteristics.}
    \vspace{-1mm}
    \label{fig:3dvis}
\end{figure}

\begin{figure}[t]
    \centering
    \includegraphics[width=1\linewidth]{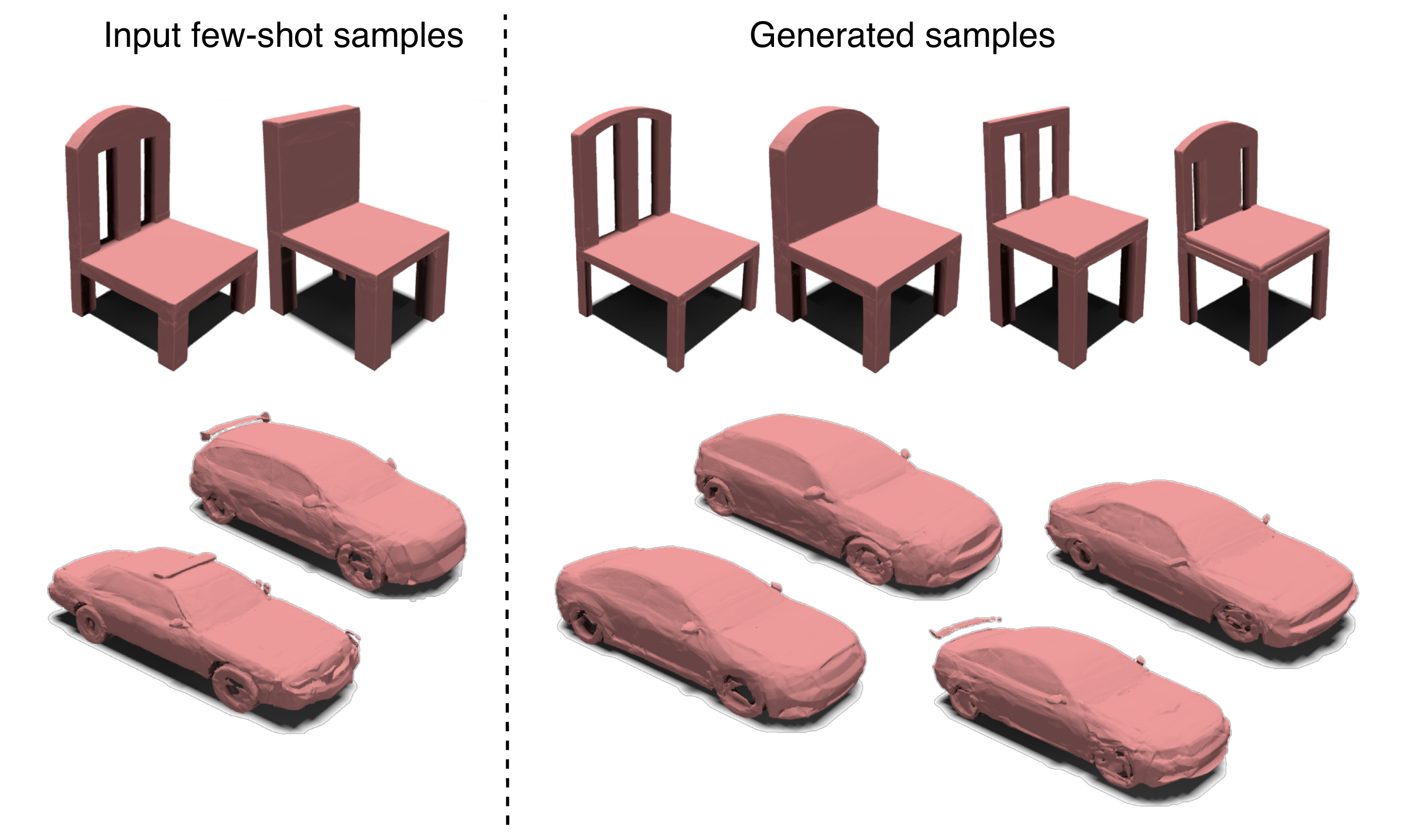}
    \vspace{-3mm}
    \caption{Visualizations of generated ShapeNet-INRs: few-shot examples and generated samples for both chair and car categories.}
    \vspace{-4mm}
    \label{fig:3dvis2}
\end{figure}

\subsection{Main Results}

The empirical results illustrated in \cref{fig:3d} and \cref{fig:2d} demonstrate the consistent superior performance of our method across all evaluation metrics in the 10-shot setting. In \cref{fig:2d}, our comprehensive evaluation on both MNIST-INRs and CIFAR10-INRs reveals that our approach surpasses both modality-based and INR-based methods in terms of quality and diversity metrics. Notably, while most existing few-shot generation frameworks address style transfer, our method is designed to tackle the more challenging task of cross-class knowledge transfer. Similar evaluation results are observed in the 3D domain, as demonstrated in \cref{fig:3d}. Qualitative results are shown in \cref{fig:3dvis} and \cref{fig:3dvis2}. There are various differences between input INR and output INR's rendered results, indicating the effectiveness of our method.


\subsection{Indepth Analysis}

\begin{figure}[tbp]
    \centering 

    \subfigbottomskip=2pt
    \subfigcapskip=-5pt 
    \subfigure[w/o smooth augmentation]{
    \includegraphics[width=0.48\linewidth]{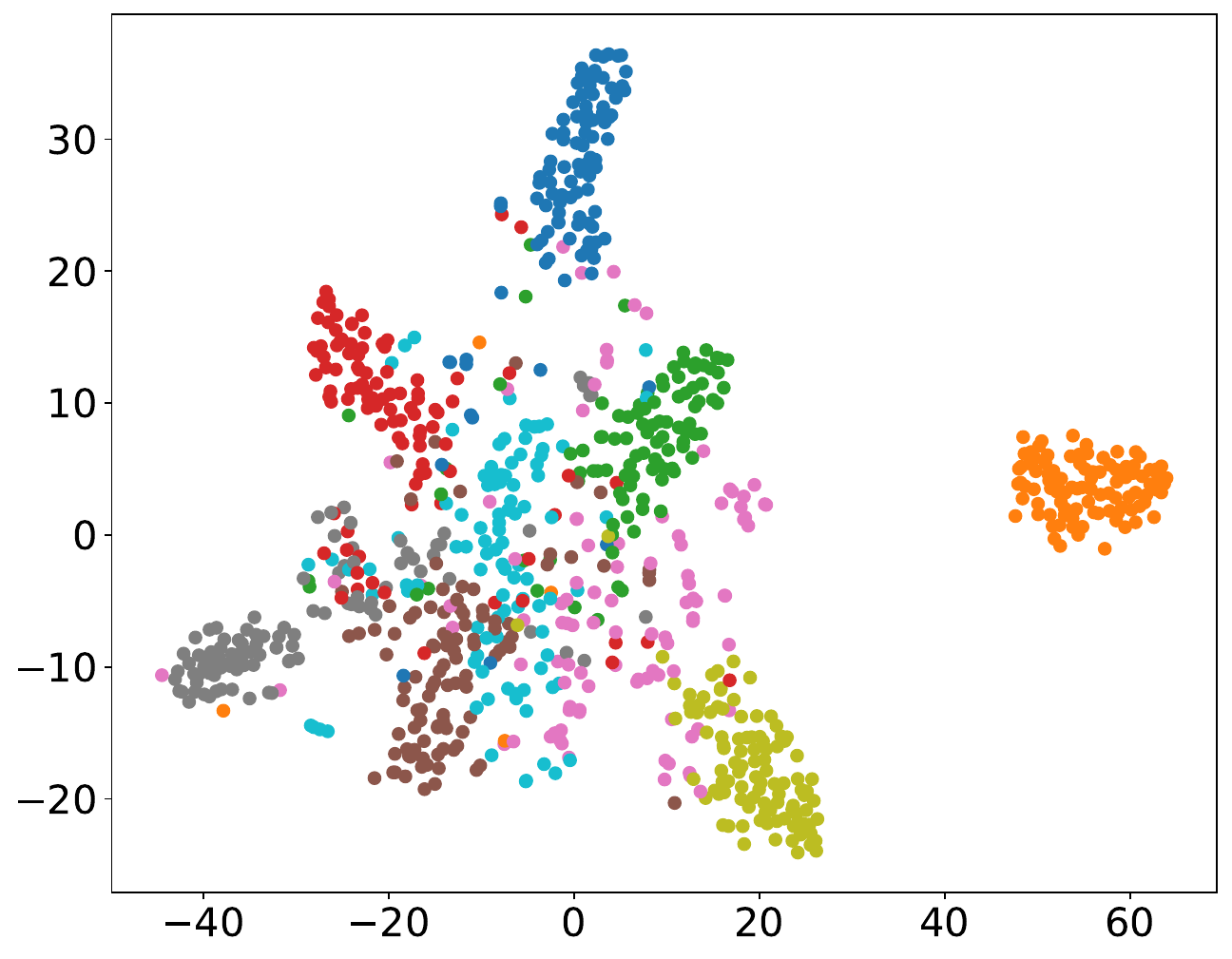}}
    \subfigure[w/ smooth augmentation]{
    \includegraphics[width=0.48\linewidth]{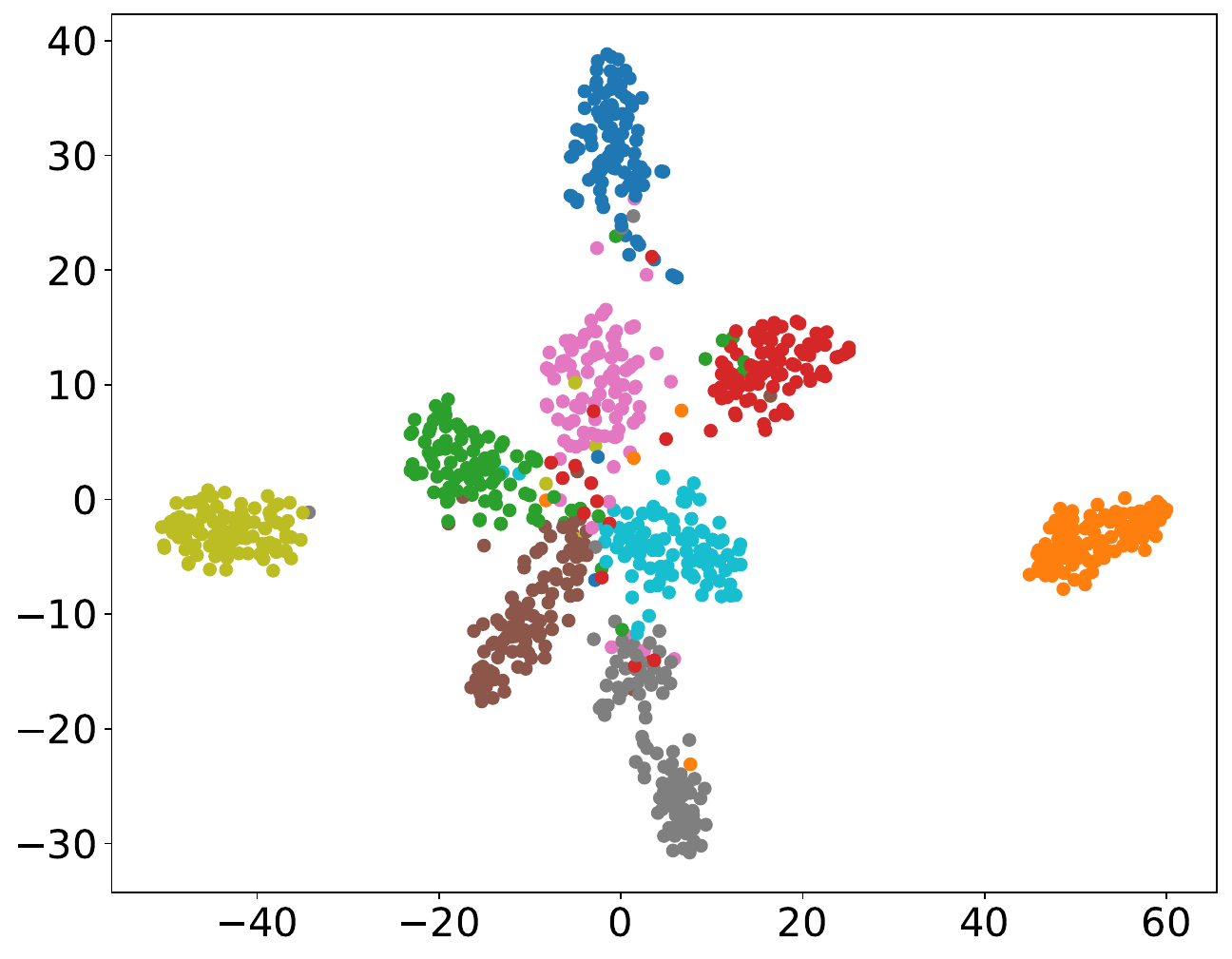}}
    \vspace{-2mm}
    \caption{2D t-SNE visualization of equivariant subspace. Smooth augmentation (right) produces more compact and discriminative category clusters compared to baseline (left), demonstrating enhanced equivariant feature learning.}
    \label{fig:tsne}
    \vspace{-2mm}
\end{figure}

\noindent\textbf{Exploration of equivariant subspace.}
The efficacy of our proposed method is fundamentally rooted in the expressive power of the equivariant subspace, which enables the incorporation of category-specific knowledge into distribution modeling. To illustrate the learning dynamics within the weight space, we conduct a 2D t-SNE visualization of the equivariant subspace, as shown in \cref{fig:tsne}. We present comparative visualizations both with and without smooth augmentation of the original weight space. The results demonstrate that the pre-trained equivariant encoder successfully projects the original weight space into a discriminative manifold where categories exhibit clear clustering behavior. Furthermore, the application of smooth augmentation to the original space alleviates the complexity of equivariant subspace learning, resulting in more compact and category-specific embeddings.


\begin{figure}[tbp]

    \centering 
    \subfigbottomskip=2pt
    \subfigcapskip=-5pt 
    \subfigure[COV with different $\gamma$]{
    \includegraphics[width=0.48\linewidth]{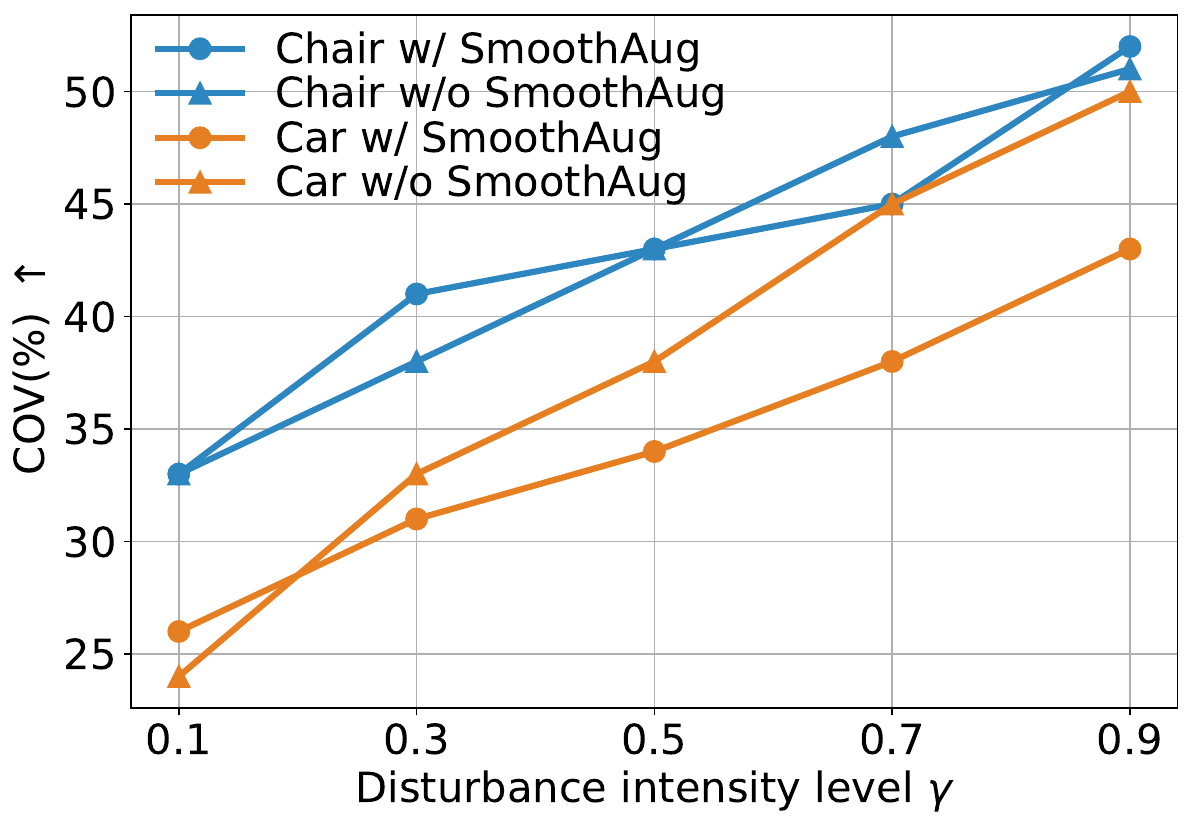}\label{fig:noise1}}
    \subfigure[MMD with different $\gamma$]{
    \includegraphics[width=0.48\linewidth]{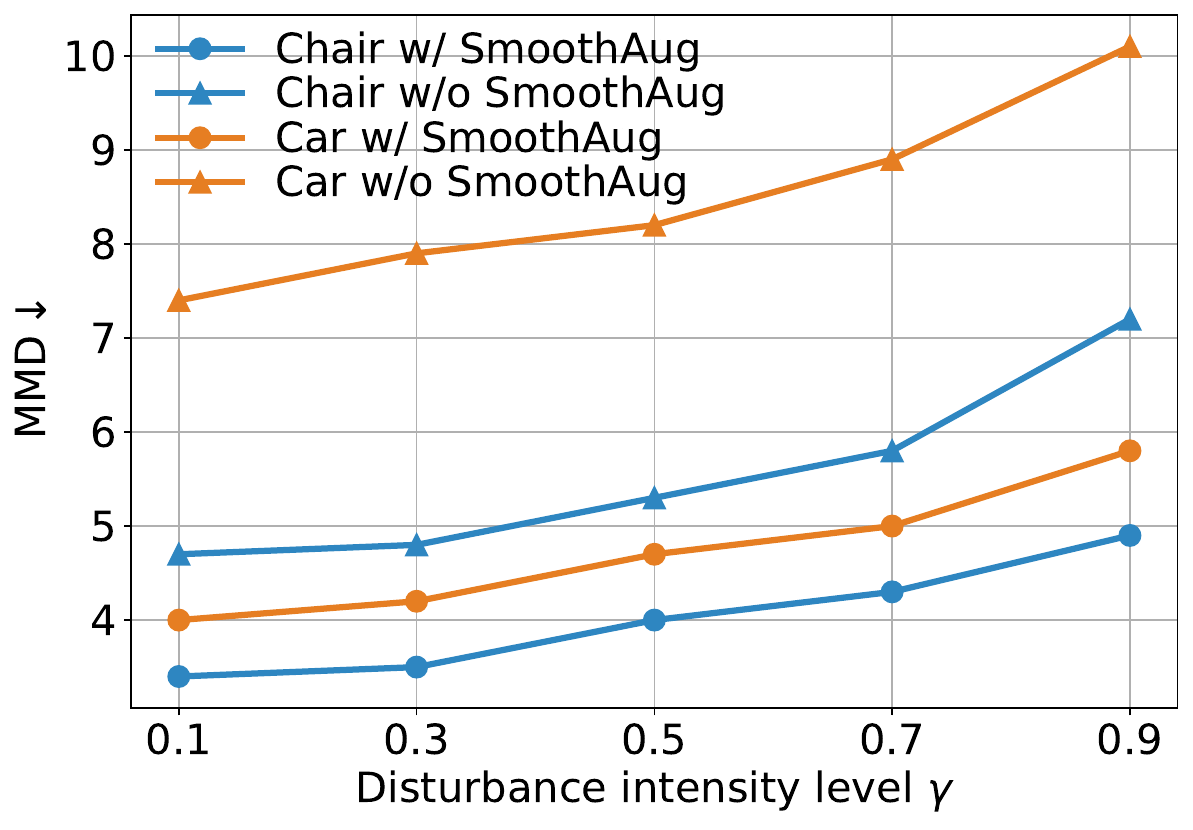}\label{fig:noise2}}
    \vspace{-2mm}
    \caption{The evaluation on ShapeNet-INRs chair and car categories with respect to different subspace disturbance intensity. (a) Higher disturbance leads to increased COV, indicating greater sample diversity. (b) However, larger disturbances result in higher MMD, reflecting decreased generation quality.}
    \vspace{-3mm}
    \label{fig:noise}
\end{figure}

\noindent\textbf{Impact of subspace disturbance.}
We conduct a systematic analysis of equivariant subspace disturbance intensity and its impact on generation performance, as illustrated in \cref{fig:noise}. Our findings demonstrate a clear trade-off between sample diversity and functional fidelity. The intensity of subspace disturbance exhibits a direct correlation with final generation variance. As shown in \cref{fig:noise1}, higher disturbance intensities lead to increased COV, indicating greater diversity among generated samples. However, \cref{fig:noise2} reveals that this enhanced diversity comes at a cost to generation quality, as measured by increasing MMD.


\noindent\textbf{Ablation study.}
To evaluate the importance of equivariance, we conduct ablation studies comparing three variants: (1) our full \algoname framework, (2) a variant that omits the equivariant encoder and uses unconditional diffusion, and (3) a variant that replaces equivariant features with class-label conditioning. As shown in \cref{fig:ab2d}, the framework with equivariance modules demonstrates superior performance in both generation quality and diversity. The unconditional variant exhibits mode collapse, and class-label conditioning variant offers only minimal improvement, demonstrating the limitations of direct class conditioning.


Another more fine-grained ablation study is conducted to discern the individual contributions of weight space smooth augmentation and equivariant subspace disturbance. The results are presented in \cref{fig:ab3d}. Our analysis reveals that smooth augmentation alone enhances both generation quality and diversity. This empirical evidence supports the idea that starting from an optimized start point, the equivariant encoder could gain better expressive power through contrastive learning.
Conversely, applying subspace disturbance yields increased diversity, although a modest decrease in generation quality appears when smooth augmentation is absent. 
This trade-off can be attributed to the fact that equivariant features with same disturbance may locate at different category clusters with and without the smooth augmentation. The simultaneous employment of both smooth augmentation and subspace disturbance enables our method to achieve optimal performance.


\begin{table}[]
\caption{Ablation study of the equivariant encoder on MNIST-INRs. The incorporation of equivariance largely enhances the overall generation quality and diversity compared to different variants that omit equivariance.}
\label{fig:ab2d}
\footnotesize
\centering
\begin{tabular}{c|cc}
\toprule
Methods & FID$\downarrow$ & LPIPS$\uparrow$ \\ \midrule
\begin{tabular}[c]{c} \algoname w/o  condition \end{tabular} & 164.78 & 0.2432 \\
\begin{tabular}[c]{c} \algoname w/ label condition \end{tabular} & 162.56 & 0.2437 \\
\begin{tabular}[c]{c} \algoname w/ equivariant condition\end{tabular} & \textbf{121.24} & \textbf{0.4133}   \\ \bottomrule
\end{tabular}%
\end{table}

\begin{table}[]

\caption{Ablation study of weight space smooth augmentation and equivariant subspace disturbance on ShapeNet-INRs, where each test set consists of INRs
from a single unseen object class.}
\label{fig:ab3d}
\resizebox{\columnwidth}{!}{%
\begin{tabular}{c|cc|ccc}
\toprule
Category          & Smooth &   Disturbance     & \multicolumn{1}{c}{MMD$\downarrow$} & \multicolumn{1}{c}{COV(\%)$\uparrow$} & \multicolumn{1}{c}{1-NNA(\%)$\downarrow$} \\ \midrule
\multirow{4}{*}{Airplane} &\textcolor{purple}{$\usym{2717}$} & \textcolor{purple}{$\usym{2717}$}   & 5.5    & 29    & 79.4      \\
 &\textcolor{teal}{$\usym{2713}$} &      \textcolor{purple}{$\usym{2717}$}      & \textbf{3.2} & 30 & 77.5 \\
 &\textcolor{purple}{$\usym{2717}$} & \textcolor{teal}{$\usym{2713}$}     &  5.7                    &      \textbf{35}                &       \underline{75.2}               \\
 & \textcolor{teal}{$\usym{2713}$}& \textcolor{teal}{$\usym{2713}$}& \underline{3.4}   &  \textbf{35}    &  \textbf{73.0}             \\ \midrule
\multirow{4}{*}{Car}      & \textcolor{purple}{$\usym{2717}$}& \textcolor{purple}{$\usym{2717}$}& 4.7   & 22    & 83.9      \\
 & \textcolor{teal}{$\usym{2713}$}&     \textcolor{purple}{$\usym{2717}$}       & \textbf{3.4} & 25 & \underline{78.4} \\
 &\textcolor{purple}{$\usym{2717}$} &    \textcolor{teal}{$\usym{2713}$}        &           4.8  &    \textbf{33}   &    78.5        \\
 & \textcolor{teal}{$\usym{2713}$}& \textcolor{teal}{$\usym{2713}$}&    \underline{3.5}  &  \underline{31}        &  \textbf{76.5}            \\ \midrule
\multirow{4}{*}{Chair}    &\textcolor{purple}{$\usym{2717}$} & \textcolor{purple}{$\usym{2717}$}& 7.4    & 30    & 77.5      \\
 &\textcolor{teal}{$\usym{2713}$} &    \textcolor{purple}{$\usym{2717}$}        & \textbf{4.2} & 30 & \underline{76.3} \\
 & \textcolor{purple}{$\usym{2717}$}&    \textcolor{teal}{$\usym{2713}$}        &          7.9 & \underline{38}   &  77.0 \\
 &\textcolor{teal}{$\usym{2713}$} & \textcolor{teal}{$\usym{2713}$} &           \textbf{4.2}        &       \textbf{41}               &        \textbf{67.1}              \\ \bottomrule
\end{tabular}%
}
\vspace{-2mm}
\end{table}

\vspace{-2mm}
\section{Conclusion}

In this work, we present a practical setting named Few-shot Implicit Function Generation and introduce solution to it by systematically leveraging the principle of equivariance. Our three-stage framework effectively addresses both the generalization and mode collapse challenges inherent in weight space generation. Extensive experiments across INR data for 2D image and 3D shape domains validate our approach, demonstrating superior performance in both generation quality and diversity.

\newpage
{
    \small
    \bibliographystyle{ieeenat_fullname}
    \bibliography{main}
}
\end{document}